# Perfect Tree-like Markovian Distributions

**Ann Becker**
Computer Science Department
Technion, Haifa, 32000, Israel
anyuta@cs.technion.ac.il

**Dan Geiger**[*]
Computer Science Department
Technion, Haifa, 32000, Israel
dang@cs.technion.ac.il

**Christopher Meek**
Microsoft Research
Redmond, WA, 98052, USA
meek@microsoft.com

## Abstract

We show that if a strictly positive joint probability distribution for a set of binary random variables factors according to a tree, then vertex separation represents all and only the independence relations encoded in the distribution. The same result is shown to hold also for multivariate strictly positive normal distributions. Our proof uses a new property of conditional independence that holds for these two classes of probability distributions.

## 1 Introduction

A useful approach to multivariate statistical modeling is to first define the conditional independence constraints that are likely to hold in a domain, and then to restrict the analysis to probability distributions that satisfy these constraints. An increasingly popular way of specifying independence constraints are directed and undirected graphical models where independence constraints are encoded through the topological properties of the corresponding graphs (Lauritzen 1982; Lauritzen and Spiegelhalter, 1988; Pearl, 1988; Whittaker, 1990).

The key idea behind these specification schemes is to utilize the correspondence between *vertex separation* in graphs and *conditional independence* in probability; each vertex represents a variable and if a set of vertices $Z$ blocks all the paths between two vertices, then the corresponding two variables are asserted to be conditionally independent given the variables corresponding to $Z$. The success of graphical models stems in part from the fact that vertex separation and conditional independence share key properties which render graphs an effective language for specifying independence constraints.

In this paper we show that when graphical models are trees and distributions are from specific classes, then the relationship between vertex separation and conditional independence is much more pronounced. More specifically, we show that if a strictly positive joint probability distribution for a set of binary random variables factors according to a tree, then vertex separation represents all and only the independence relations encoded in the distribution. The same result is shown to hold also for multivariate strictly positive normal distributions.

The class of Markov trees has been studied in several contexts. Practical algorithms for learning Markov trees from data have been used for pattern recognition (Chow and Liu, 1968). Geometrical properties of families of tree-like distributions have been studied in (Settimi and Smith, 1999). Finally, the property of perfectness, when a graphical model represents all and only the conditional independence facts encoded in a distribution, is a key assumption in learning causal relationships from observational data (Glymour and Cooper, 1999).

## 2 Preliminaries

Throughout this article we use lowercase letters for single random variables (e.g., $x, y, z$) and boldfaced lowercase letters (e.g., $\mathbf{x}, \mathbf{y}, \mathbf{z}$) for specific values for these random variables. Set of random variables are denoted by capital letters (e.g., $X, Y, Z$), and their values are denoted by boldfaced capital letters (e.g., $\mathbf{X}, \mathbf{Y}, \mathbf{Z}$). For example, if $Z = \{x, y\}$ then $\mathbf{Z}$ stands for $\{\mathbf{x}, \mathbf{y}\}$ where $\mathbf{x}$ is a value of $x$ and $\mathbf{y}$ is a value of $y$. We use $P(\mathbf{X})$ as a short hand notation for $P(X = \mathbf{X})$. We say that $P(\mathbf{X})$ is *strictly positive* if $\forall \mathbf{X}\ P(\mathbf{X}) > 0$. We use $Xy$ as a short hand notation for $X \cup \{y\}$.

Let $X, Y$ and $Z$ be three disjoint sets of random variables having a joint probability distribution $P(X, Y, Z)$. Then, $X$ and $Y$ are conditionally independent given $Z$, denoted by $X \perp_P Y \mid Z$, if and only if

$$\forall \mathbf{X} \forall \mathbf{Y} \forall \mathbf{Z}\ \ P(\mathbf{X}, \mathbf{Y}, \mathbf{Z})P(\mathbf{Z}) = P(\mathbf{X}, \mathbf{Z})P(\mathbf{Y}, \mathbf{Z}).$$

When $P$ is strictly positive an equivalent definition is that $X \perp_P Y \mid Z$ holds if and only if

$$\forall \mathbf{X} \forall \mathbf{Y} \forall \mathbf{Z}\ \ P(\mathbf{X}|\mathbf{Z}) = P(\mathbf{X}|\mathbf{Y}, \mathbf{Z}).$$

When $P(X, Y, Z)$ is a strictly positive joint normal distribution, then $X$ and $Y$ are conditionally independent given $Z$ if and only if $\rho_{xy \cdot Z} = 0$ for every $x \in X$ and $y \in Y$ where $\rho_{xy.Z}$ is the partial correlation coefficient of $x$ and $y$ given $Z$ (Cramer, 1946).

---

[*]Part of this work was done while the author was on sabbatical at Microsoft research.

The ternary relation $X \perp_P Y \mid Z$ was introduced in (Dawid, 1979) and further studied in (e.g., Spohn 1980; Pearl and Paz 1987; Pearl 1988; Geiger and Pearl 1993; Studeny 1992). The ternary relation $X \perp_P Y \mid Z$ satisfies the following five properties which are called the graphoid axioms (Pearl and Paz, 1987).

- **Symmetry:**
$$X \perp_P Y \mid Z \Rightarrow Y \perp_P X \mid Z \quad (1)$$

- **Decomposition:**
$$X \perp_P YW \mid Z \Rightarrow X \perp_P Y \mid Z \quad (2)$$

- **Weak Union:**
$$X \perp_P YW \mid Z \Rightarrow X \perp_P Y \mid ZW \quad (3)$$

- **Contraction:**
$$X \perp_P Y \mid Z \wedge X \perp_P W \mid ZY \Rightarrow$$
$$X \perp_P YW \mid Z \quad (4)$$

If $P$ is strictly positive, then

- **Intersection:**
$$X \perp_P Y \mid ZW \wedge X \perp_P W \mid ZY \Rightarrow$$
$$X \perp_P YW \mid Z \quad (5)$$

The following property holds for joint normal distributions $P(X, Y, Z, c)$ (Pearl, 1988). It also holds for discrete random variables if $Z = \emptyset$ and $c$ is a binary random variable.

- **Weak Transitivity:**
$$X \perp_P Y \mid Z \wedge X \perp_P Y \mid Zc \Rightarrow$$
$$X \perp_P c \mid Z \vee c \perp_P Y \mid Z \quad (6)$$

A *Markov network* of a probability distribution $P(x_1, \ldots, x_n)$ is an undirected graph $G = (V, E)$ where $V = \{x_1, \ldots, x_n\}$ is a set of vertices, one for each random variable $x_i$, and $E$ is a set of edges each represented as $(x_i, x_j)$ such that $(x_i, x_j) \in E$ if and only if
$$\neg x_i \perp_P x_j \mid \{x_1, \ldots, x_n\} \setminus \{x_i, x_j\}.$$
A *Markov tree* is a Markov network where $G$ is a tree.

A key property of Markov networks is the following. Let $A \perp_G B \mid C$ stand for the assertion that every path in $G$ between a vertex in $A$ and a vertex in $B$ passes through a vertex in $C$, where $A, B$, and $C$ are mutually disjoint sets of vertices. Note that whenever $A \perp_G B \mid C$ holds in $G$, $A$ and $B$ are (vertex) separated by $C$. The ternary relation $A \perp_G B \mid C$ satisfies all the properties we listed for $A \perp_P B \mid C$ and some additional properties that do not hold for $A \perp_P B \mid C$ (Pearl, 1988).

**Theorem 1 (Pearl and Paz, 1987; Pearl, 88)**
Let $G$ be a Markov network of $P(x_1, \ldots, x_n)$, and suppose Intersection holds for $P$. Then
$$A \perp_G B \mid C \text{ implies } A \perp_P B \mid C \quad (7)$$
for every disjoint set of vertices $A, B,$ and $C$ of $G$ and their corresponding random variables in $\{x_1, \ldots, x_n\}$.

The main result in this paper is a converse to Eq. 7 under suitable conditions. When the converse holds we say that $G$ is a *perfect representation* of $P$. To facilitate our argument we must first introduce a new property for conditional independence.

- **Decomposable transitivity:**
$$aB \perp_P De \mid c \wedge a \perp_P e \mid BD \Rightarrow$$
$$a \perp_P c \mid B \vee c \perp_P e \mid D \quad (8)$$

## 3 New property of conditional independence

We now prove that decomposable transitivity holds for strictly positive joint probability distributions of binary random variables and for strictly positive normal distributions. We then show that decomposable transitivity holds also for vertex separation in undirected graphs.

**Theorem 2** *Let $a, c, e$ be binary random variables, $B$ and $D$ be (possibly empty) sets of binary random variables, and $P(a, c, e, B, D)$ be a strictly-positive joint probability distribution for these random variables. Then*
$$aB \perp_P De \mid c \wedge a \perp_P e \mid BD \Rightarrow$$
$$a \perp_P c \mid B \vee c \perp_P e \mid D$$
*holds for $P$.*

**Proof:** We use **a** to denote a value for $a$, **B** to denote a value for a set of variables $B$, and $a^0$ and $a^1$ to denote the two values of a binary random variable $a$.

Due to $aB \perp_P De \mid c$ it follows that
$$P(\mathbf{a}, \mathbf{B}, \mathbf{c}, \mathbf{D}, \mathbf{e}) \cdot P(\mathbf{c}) = P(\mathbf{a}, \mathbf{B}, \mathbf{c}) \cdot P(\mathbf{c}, \mathbf{D}, \mathbf{e}) \quad (9)$$
for every value $\mathbf{a}, c, e, B, D$ of the corresponding random variables. Due to $a \perp_P e \mid BD$ it follows that
$$P(\mathbf{a}^0, \mathbf{B}, \mathbf{D}, \mathbf{e}^0) \cdot P(\mathbf{a}^1, \mathbf{B}, \mathbf{D}, \mathbf{e}^1) =$$
$$P(\mathbf{a}^1, \mathbf{B}, \mathbf{D}, \mathbf{e}^0) \cdot P(\mathbf{a}^0, \mathbf{B}, \mathbf{D}, \mathbf{e}^1). \quad (10)$$
for every value $\mathbf{B}, \mathbf{D}$ of $B, D$. Since $c$ is a binary variable
$$P(\mathbf{a}, \mathbf{B}, \mathbf{D}, \mathbf{e}) = P(\mathbf{a}, \mathbf{B}, \mathbf{c}^0, \mathbf{D}, \mathbf{e}) + P(\mathbf{a}, \mathbf{B}, \mathbf{c}^1, \mathbf{D}, \mathbf{e}) \quad (11)$$

Now, substituting Eq. 9 into Eq. 11, then substituting the result into Eq. 10, yields using some divisions, which are allowed because $P$ is strictly positive, that
$$1 + \alpha(\mathbf{B})\beta(\mathbf{D}) = \alpha(\mathbf{B}) + \beta(\mathbf{D})$$
where
$$\alpha(\mathbf{B}) = \frac{P(\mathbf{a}^1, \mathbf{B}, \mathbf{c}^0) \cdot P(\mathbf{a}^0, \mathbf{B}, \mathbf{c}^1)}{P(\mathbf{a}^0, \mathbf{B}, \mathbf{c}^0) \cdot P(\mathbf{a}^1, \mathbf{B}, \mathbf{c}^1)}$$
and
$$\beta(\mathbf{D}) = \frac{P(\mathbf{c}^1, \mathbf{D}, \mathbf{e}^0) \cdot P(\mathbf{c}^0, \mathbf{D}, \mathbf{e}^1)}{P(\mathbf{c}^0, \mathbf{D}, \mathbf{e}^0) \cdot P(\mathbf{c}^1, \mathbf{D}, \mathbf{e}^1)}$$

Consequently, either $\alpha(\mathbf{B}) = 1$ or $\beta(\mathbf{D}) = 1$. Furthermore, since $\mathbf{B}$ and $\mathbf{D}$ are arbitrary values of $B$ and $D$, respectively, we have $\forall \mathbf{B} \forall \mathbf{D} [\alpha(\mathbf{B}) = 1 \vee \beta(\mathbf{D}) = 1]$ which is equivalent to $[\forall \mathbf{B}\ \alpha(\mathbf{B}) = 1] \vee [\forall \mathbf{D}\ \beta(\mathbf{D}) = 1]$ which is equivalent to $a \perp_P c \mid B \vee c \perp_P e \mid D$. $\square$

**Theorem 3** *Let $a, c$, and $e$ be continuous random variables, $B$ and $D$ be (possibly empty) sets of continuous random variables, and let $P(a, c, e, B, D)$ be a strictly positive joint normal probability distribution for these random variables. Then,*

$$aB \perp_P De \,|\, c \;\wedge\; a \perp_P e \,|\, BD \;\Rightarrow$$
$$a \perp_P c \,|\, B \;\vee\; c \perp_P e \,|\, D \quad (12)$$

*holds for $P$.*

**Proof:** We use a formal logical deduction style to emphasize that the only properties of normal distributions being used are the ones encoded in Symmetry, Decomposition, Intersection, Weak union, and Weak transitivity. Recall that weak transitivity holds for every normal distribution and that intersection holds for strictly positive normal distributions. The other properties hold for every probability distribution.

We now derive the conclusion of Eq. 12 from its antecedents.

1. $aB \perp_P De \,|\, c$
   (Given)
2. $a \perp_P e \,|\, BD$
   (Given)
3. $a \perp_P D \,|\, cB$
   (W. union, Decomposition, and Symmetry on (1))
4. $B \perp_P e \,|\, cD$
   (W. union, Decomposition, and Symmetry on (1))
5. $a \perp_P e \,|\, BDc$
   (Weak union and Symmetry on (1))
6. $a \perp_P c \,|\, BD \vee c \perp_P e \,|\, BD$
   (Weak transitivity on (2) and (5))
7. $a \perp_P cD \,|\, B \vee Bc \perp_P e \,|\, D$
   (Intersection and Symmetry on (3), (4) and (6))
8. $a \perp_P c \,|\, B \vee c \perp_P e \,|\, D$
   (Symmetry and Decomposition on (7))

$\square$

**Theorem 4** *Let $a, c$, and $e$ be distinct vertices of an undirected graph $G$, and let $B$ and $D$ be two (possibly empty) disjoint sets of vertices of $G$ that do not include $a, c$ or $d$. Then,*

$$aB \perp_G De \,|\, c \;\wedge\; a \perp_G e \,|\, BD \;\Rightarrow$$
$$a \perp_G c \,|\, B \;\vee\; c \perp_G e \,|\, D \quad (13)$$

*holds for $G$.*

**Proof:** Assume the conclusion of Eq. 13 does not hold in $G$ but its antecedents hold. Then, there exists a path $\gamma_1$ in $G$ between $a$ and $c$ such that no vertices from $B$ reside on $\gamma_1$, and there exists a path $\gamma_2$ in $G$ between $c$ and $e$ such that no vertices from $D$ reside on $\gamma_1$. If $B$ and $D$ are empty, then the concatenated path $\gamma_1\gamma_2$ contradicts $a \perp_G e \,|\, BD$ which is assumed to hold in $G$. Thus, we can assume either $B$ or $D$ are not empty. The concatenated path $\gamma_1\gamma_2$ contains a vertex from $B$ or $D$ (or both) because $a \perp_G e \,|\, BD$ is assumed to hold in $G$. Assume a vertex $d \in D$ resides on the path $\gamma_1$ between $a$ and $c$, or that a vertex $b \in B$ resides on the path $\gamma_2$ between $c$ and $e$. In the first case vertices $a$ and $d$ are connected and the path that connects them does not include $c$, and in the second case vertex $b$ and $e$ are connected and the path that connects them does not include $c$. Thus, in both cases, $aB \perp_G De \,|\, c$ does not hold in $G$, contradicting our assumption. $\square$

## 4　Perfect Markovian trees

We are ready to prove the main result.

**Theorem 5** *Let $G$ be a Markov tree for a probability distribution $P(x_1, \ldots, x_n)$. If $x_1, \ldots, x_n$ are binary random variables and $P$ is a strictly-positive joint probability distribution, or if $x_1, \ldots, x_n$ are continuous random variables and $P$ is a strictly positive joint normal distribution then, in both cases,*

$$A \perp_G B \,|\, C \quad \text{if and only if} \quad A \perp_P B \,|\, C \quad (14)$$

*for every disjoint set of vertices $A$, $B$, and $C$ of $G$ and their corresponding random variables in $\{x_1, \ldots, x_n\}$.*

**Proof:** Theorem 1 proves one direction of Eq. 14, and so it remains to prove that

$$A \perp_P B \,|\, C \quad \text{implies} \quad A \perp_G B \,|\, C \quad (15)$$

To prove Eq. 15 it is sufficient to show that

$$a \perp_P b \,|\, C \quad \text{implies} \quad a \perp_G b \,|\, C$$

for every pair of vertices $a \in A$ and $b \in B$ because $A \perp_P B \,|\, C$ implies $\forall a \forall b \; a \perp_P b \,|\, C$ and $\forall a \forall b \; a \perp_G b \,|\, C$ is equivalent by definition to $A \perp_G B \,|\, C$.

We proceed by contradiction. Let $x$ and $y$ be a pair of vertices for which there exists a set of vertices $Z$ satisfying

$$x \perp_P y \,|\, Z \wedge \neg x \perp_G y \,|\, Z \quad (16)$$

and such that $x$ and $y$ are connected with the shortest path among all pairs $x', y'$ for which there exists a set $Z'$ satisfying $x' \perp_P y' \,|\, Z' \wedge \neg x' \perp_G y' \,|\, Z'$.

Suppose first that the path between $x$ and $y$ is merely an edge connecting the two vertices. We will now reach a contradiction by showing that $G$ cannot be a Markov network of $P$. In particular, we show that $P$ satisfies $x \perp_P y \,|\, U_{xy}$ where $U_{xy}$ are all vertices except $x$ and $y$. Let $U_x$ be all the vertices on the $x$ side of the edge $(x, y)$ and $U_y$ be the rest of the vertices. (Namely, $U_x$ are the vertices in the component of $x$ after removing the edge $(x, y)$). Let $B = U_x \cap Z$ and $D = U_y \cap Z$. We proceed by a formal deduction using properties of conditional independence.

1. $B \perp_G Dy \,|\, x$
   (By definition of $B$ and $D$ in $G$)
2. $B \perp_P Dy \,|\, x$
   (From (1) and since $G$ is a Markov network of $P$)
3. $B \perp_P y \,|\, xD$
   (Weak union on (2))
4. $x \perp_P y \,|\, BD$
   ($Z = BD$ and $x \perp_P y \,|\, Z$ is assumed)

5. $xB \perp_P y \mid D$
   (Intersection and Symmetry on (3) and (4))
6. $x \perp_P y \mid D$
   (Decomposition on (5))
7. $x \perp_G D \mid y$
   (By definition of $D$ in $G$)
8. $x \perp_P D \mid y$
   (From (7) and since $G$ is a Markov network of $P$)
9. $x \perp_P yD \mid \emptyset$
   (Intersection on (6) and (8))
10. $x \perp_P y \mid \emptyset$
    (Decomposition on (9))
11. $x \perp_G U_y \mid y$
    (Definition of $U_y$)
12. $x \perp_P U_y \mid y$
    (From (11) and since $G$ is a Markov network of $P$)
13. $x \perp_P yU_y \mid \emptyset$
    (Contraction on (10) and (12))
14. $U_x \perp_G yU_y \mid x$
    (Definition of $U_x$ and $U_y$)
15. $U_x \perp_P yU_y \mid x$
    (From (14) and since $G$ is a Markov network of $P$)
16. $xU_x \perp_P yU_y \mid \emptyset$
    (Contraction and Symmetry on (13) and (15))
17. $x \perp_P y \mid U_xU_y$
    (Weak union and Symmetry on (16))

Now suppose the path between $x$ and $y$ has more than one edge and that $c$ is a vertex on this path. We reach a contradiction by showing that the pair $x, y$ is not the closest pair of vertices that satisfy Eq. 16 for some set $Z'$, contrary to our selection of these vertices. Let $B, D$ be a partition of $Z$ such that $B$ are the vertices in $Z$ on the $x$ side of $c$ and $D = Z \setminus B$. The rest of the derivation is a formal deduction using properties of conditional independence.

1. $xB \perp_G Dy \mid c$
   (By definition of $B$ and $D$ in $G$)
2. $xB \perp_P Dy \mid c$
   (From (1) and since $G$ is a Markov network of $P$)
3. $x \perp_P y \mid BD$
   ($Z = BD$ and $x \perp_P y \mid Z$ is assumed)
4. $x \perp_P c \mid B \vee c \perp_P y \mid D$
   (Decomposable transitivity on (2) and (3))
5. $\neg x \perp_G c \mid B \wedge \neg c \perp_P y \mid D$
   (By definition of $B$ and $D$ in $G$)
6. $[x \perp_P c \mid B \wedge \neg x \perp_G c \mid B]$ ∨
   $[c \perp_P y \mid D \wedge \neg c \perp_G y \mid D]$    ((4) and (5))

Each disjunct in Step (6) exhibits a pair of vertices that are closer to each other than $x$ and $y$ and yet satisfy Eq. 16 for some set $Z'$. Note that Step (4) uses Decomposable transitivity which holds if $x_1, \ldots, x_n$ are binary random variables and $P$ is a strictly-positive joint probability distribution, or if $x_1, \ldots, x_n$ are continuous random variables and $P$ is a strictly positive joint normal distribution, as assumed. □

## 5 Discussion

Our proof uses a new property of conditional independence that holds for the two classes of probability distributions we have focused on. The approach of using logical properties of conditional independence as a way of reasoning follows the approach taken by (Pearl and Paz, 1987) who analyzed the logical properties shared by vertex separation and conditional independence.

The algorithmic consequence of Theorem 5 is that in order to check whether a Markov tree of $P$ represents all the conditional independence statements that hold in $P$, assuming $P$ satisfies Intersection and Decomposable transitivity, requires one to check whether for each edge $(x, y)$ in $G$, $x \perp_P y \mid \emptyset$ holds in $P$. Note that this test is more reliable and simpler than checking for each edge $(x, y)$ in $G$, whether $x \perp_P y \mid U_{xy}$ holds, as the definition of a Markov tree requires. An open question remains as to what is the minimal computation needed to ensure that a general Markov network represents all the conditional independence statements that hold in $P$ and what properties $P$ needs to satisfy to accommodate these computations.

A straightforward attempt to extend our results without changing the tests or the assumptions on $P$ is quite limited because we have counter examples to Theorem 5 when $G$ is a polytree (a directed graph with no underlying undirected cycles) and when $P$ does not satisfy Intersection or Decomposable transitivity. These counter examples, together with the proof of Theorem 5, show that if $G$ is a Markov tree of a probability distribution $P$, then $G$ is a perfect representation of $P$ if and only if $P$ satisfies Intersection and Decomposable transitivity.

## References

Chow, C., & Liu, C. (1968). Approximating discrete probability distributions with dependence trees. *IEEE Transaction on information theory, 14*, 462–467.

Cramer, H. (1946). *Mathematical methods of statistics*. Princeton university press.

Dawid, A. P. (1979). Conditional independence in statistical theory (with discussion). *Journal of the royal statistical society, Series B, 41*, 1–31.

Geiger, D., & Pearl, J. (1993). Logical and algorithmic properties of conditional independence and graphical models. *Annals of Statistics, 21*, 2001–21.

Glymour, C., & Cooper, G. (1999). *Computation, Causation, Discovery*. AAAI press/The MIT press, Menlo Park.

Lauritzen, S. (1989). *Lectures on contingency tables (3rd edition)*. Aalborg University.

Lauritzen, S., & Spiegelhalter, D. (1988). Local computations with probabilities on graphical structures and their application to expert systems (with discussion). *Journal Royal Statistical Society, B, 50*, 157–224.

Pearl, J. (1988). *Probabilistic reasoning in intelligent systems: Networks of plausible inference*. Morgan Kaufmann, San Mateo, California.

Pearl, J., & Paz, A. (1987). Graphoids: a graph based logic for reasoning about relevancy relations. In Boulay, B., Hogg, D., & Steel, L. (Eds.), *Advances in artificial intelligence-II*, pp. 357–363. North-Holland, Amsterdam.

Settimi, R., & Smith, J. (1999). Geometry, moments and bayesian networks with hidden variables. In *Proceedings of the seventh international workshop on Artificial Intelligence and Statitics*, pp. 293–298 San Francisco. Morgan Kaufmann.

Spohn, W. (1980). Stochastic independence, causal independence, and shieldability. *Journal philosophical logic, 9*, 73–99.

Studeny, M. (1992). Conditional independence have no finite complete characterization. In *Transactions of the 11th Prague conference on information theory, statistical decision functions and random processes*, pp. 3–16 Prague. Academia.

Whittaker, J. (1990). *Graphical models in applied multivariate statistics*. John Wiley and Sons, Chichester.